\documentclass{article}
\usepackage[utf8]{inputenc}
\usepackage{algorithm}
\usepackage{algpseudocode}
\usepackage{setspace}
\usepackage{amsmath}
\usepackage{empheq}
\usepackage{caption}
\usepackage{subcaption}
\usepackage{tikz}
 
\usepackage{PRIMEarxiv}

\usepackage{cite}
\usepackage{amsmath,amssymb,amsfonts}
\usepackage{graphicx}
\usepackage{textcomp}
\usepackage{xcolor}

\title{Input Normalized Stochastic Gradient Descent Training of Deep Neural Networks}

\author{
  Salih Atici, Hongyi Pan, Ahmet Enis Cetin \\
  Department of Electrical and Computer Engineering\\
  University of Illinois Chicago\\
  Chicago, Illinois, USA\\
  \texttt{\{hpan21, satici2, aecyy\}@uic.edu} \\
}
\date{December 2022}

\begin{document}

\maketitle

\begin{abstract}
In this paper, we propose a novel optimization algorithm for training machine learning models called Input Normalized Stochastic Gradient Descent (INSGD), inspired by the Normalized Least Mean Squares (NLMS) algorithm used in adaptive filtering. When training complex models on large datasets, the choice of optimizer parameters, particularly the learning rate, is crucial to avoid divergence. Our algorithm updates the network weights using stochastic gradient descent with $\ell_1$ and $\ell_2$-based normalizations applied to the learning rate, similar to NLMS. However, unlike existing normalization methods, we exclude the error term from the normalization process and instead normalize the update term using the input vector to the neuron. Our experiments demonstrate that our optimization algorithm achieves higher accuracy levels compared to different initialization settings. We evaluate the efficiency of our training algorithm on benchmark datasets using ResNet-18, WResNet-20, ResNet-50, and a toy neural network. Our INSGD algorithm improves the accuracy of ResNet-18 on CIFAR-10 from 92.42\% to 92.71\%, WResNet-20 on CIFAR-100 from 76.20\% to 77.39\%, and ResNet-50 on ImageNet-1K from 75.52\% to 75.67\%.
\end{abstract}

\section{Introduction}

Deep Neural Networks (DNNs) have gained immense popularity and have been extensively applied across various research fields due to their convenience and ease of use \cite{lecun1995convolutional, he2016deep, krizhevsky2017imagenet, simonyan2014very, long2015fully}. Researchers from different domains can readily utilize DNN models for their work, as these models can adapt their parameters to find optimal solutions for a wide range of problems, particularly in supervised learning scenarios. The parameters of a DNN model are updated using optimization algorithms, and researchers have proposed various algorithms that offer fresh perspectives and address different conditions \cite{ruder2016overview}. It is important to note that different optimization algorithms can yield varying results depending on the task at hand.

Among the commonly used optimization algorithms for supervised learning in DNN models, Stochastic Gradient Descent (SGD) is widely adopted and can produce remarkable results on large-scale datasets when appropriate initial conditions are set \cite{bottou2010large}. Another popular algorithm, called Adam, can outperform SGD in suitable circumstances; however, it is crucial to choose an initial learning rate carefully, as a relatively high value can lead to divergence \cite{kingma2014adam}.
While optimization algorithms play a significant role in training DNN models, ensuring convergence of weights and finding the optimal solution for a given problem is not always guaranteed. The evaluation of an optimization algorithm should also consider its limitations. This paper addresses two limitations in the convergence of Deep Neural Network (DNN) models: the impact of different choices of hyperparameters and input power. To overcome these limitations, we propose a novel optimization algorithm called Normalized Input Stochastic Gradient Descent (NISGD), which draws inspiration from the Normalized Least Mean Squares (NLMS) algorithm used in adaptive filtering. Our study focuses on demonstrating how NISGD is inspired by NLMS and how it can effectively address problems caused by various factors. 

\subsection{Stochastic Gradient Descent} \label{sgd}
Stochastic gradient descent (SGD) is an iterative optimization method commonly used in machine learning to update the weights of a network model. It calculates the gradient of the weights based on the objective function defined to measure the error in the training of the model and it estimates the new set of weights using the gradients with a predefined step size. Stochastic gradient descent is proven to converge to the optimal set of weights with the correct choice of step size and the initial settings. The gradual convergence provided by gradient descent helps us to optimize the weights used in any type of machine learning model. 

Assume a pair of $(\mathbf{x},\mathbf{y})$ composed of an arbitrary input $\mathbf{x}$ and an output $\mathbf{y}$. Given a set of weights $\mathbf{w}\in \mathbb{W}$ where $\mathbb{W}$ stands for the space of possible weights, a machine learning model predicts the output using a nonlinear function $f(\mathbf{x},\mathbf{w})$ and the optimal weights, $\mathbf{w}^*$, to minimize the objective (loss) function $L(\mathbf{y}, f(\mathbf{x},\mathbf{w}))$
\begin{equation}    
\mathbf{w}^* = \arg\min_{\mathbf{w}\in \mathbb{W}} L(\mathbf{y}, f(\mathbf{x},\mathbf{w}))\label{eq: loss}
\end{equation}

Due to the highly complex and non-linear nature of machine learning models, it is impossible to find a closed-form solution for the optimization problem given in Eq.~(\ref{eq: loss}) \cite{adapt1}. The gradient descent algorithm is introduced to avoid extensive computation and give an iterative method to estimate the optimal weights. The formula for SGD is given as:
\begin{equation}   
\mathbf{w}(k+1) = \mathbf{w}(k) + \lambda \nabla_{\mathbf{w}(k)} L(\mathbf{y}, f(\mathbf{x},\mathbf{w}))
\end{equation}
where $\mathbf{w}(j)$ represents the weights at $j^{th}$ step, $\nabla_{\mathbf{w}(k)}L $ is the gradient of the objective function with respect to the weights being updated and $\lambda$ is the step size or the learning rate.

Although Stochastic Gradient Descent (SGD) is a simple algorithm that can be applied to various tasks, it faces challenges related to tuning and scalability, which hinder its ability to converge quickly in deep learning algorithms. If the initial weights are not properly defined, and without preconditioned gradients that consider curvature information, the algorithm can become trapped in local minima \cite{le2011optimization, hinton2006reducing}. To estimate the minimum of the objective function more effectively, a deeper understanding of the error surface is required. In addition to using gradients, the exploitation of second-order derivatives can lead to faster convergence. Addressing these challenges involves considering the Hessian matrix of the objective function. However, calculating the second derivative with respect to each weight is computationally expensive and can lead to memory issues in deep networks. The Hessian matrix and its approximations are also utilized in the Normalized Least Mean Squares (NLMS) method, which will be discussed in the following subsection.

\subsection{Normalized Least Mean Squares} \label{nlms}

In adaptive filtering, assume $\mathbf{u}$, the input to a system, is a $1 \times M$ random vector with zero mean and a positive-definite covariance matrix $\mathbf{R}_\mathbf{u}$ and $d$, the desired output of the system, is a scalar random variable with zero mean and a finite variance ${\sigma}_d^2$. The linear estimation problem is defined as the solution of
\begin{equation}    
\min_\mathbf{w} \mathbb{E} \left|d - \mathbf{u} \mathbf{w}\right|^2\label{eq: linear estimation problem}
\end{equation}
where $\mathbf{w}$ is a vector containing the filter coefficients to be optimized. The linear estimation problem declares the cost function as the mean-square error and it is defined as 
\begin{equation} \label{eq:msecost_nlms}
    J(w) = \mathbb{E}  \left|d - \mathbf{u} \mathbf{w}\right|^2 = \mathbb{E} (d - \mathbf{u} \mathbf{w}) (d - \mathbf{u} \mathbf{w})^T
\end{equation}
where $(.)^T$ denotes a transpose. If we expand Eq. (\ref{eq:msecost_nlms}), it is straightforward to obtain the cost function $J(w)$ in terms of the covariance and cross-covariance matrices:
\begin{equation} \label{eq:costopen}
    J(w) = \mathbb{E} (d - \mathbf{u} \mathbf{w}) (d - \mathbf{u} \mathbf{w})^T = \sigma_d^2-\mathbf{R}_\mathbf{du}^T\mathbf{w} - \mathbf{w}^T\mathbf{R}_\mathbf{du} + \mathbf{w^T}\mathbf{R}_\mathbf{u}\mathbf{w}
\end{equation}
where $\mathbf{R}_\mathbf{du}=\mathbb{E}d\mathbf{u}$ denotes the cross-covariance vector of $d$ and $\mathbf{u}$. The closed-form solution to such a problem in (\ref{eq:costopen}) can be found using the linear estimation theory as $\mathbf{R}_\mathbf{u} w^o = \mathbf{R}_\mathbf{du}$; however, it may not be possible to obtain a closed-form solution for problems with criteria other than the mean-square-error criterion.
 
The Least Mean Squares (LMS) algorithm, developed by Widrow \textit{et al.} in the 1960s, computes the stochastic gradient and updates the weight vector iteratively to find the optimal solution for Eq.~(\ref{eq: linear estimation problem}). The optimal weight vector, denoted as $\mathbf{w}^o$, can be updated using the following iterative process.

\begin{equation}
\mathbf{w}(j) = \mathbf{w}(j-1) + \lambda \mathbf{u}^T(j) \mathbf{e}(j)\label{eq: linear estimation update}
\end{equation}
where 
$\mathbf{u} (j)$ is $j$-th observation of the random vector $\mathbf{u}$ and 
$\mathbf{e}(j)=\mathbf{d}(j)-\mathbf{u}^T\mathbf{w}(j-1)$ is the error vector at time $j$ or corresponding to the $j$-th observation of the random vector $\mathbf{u}$ and the random variable $\mathbf{d}$ \cite{widrow1960adaptive, LMS1}. The updating term is obtained as the negative of the stochastic gradient of the mean squared error function defined in Eq.~(\ref{eq: linear estimation problem}) with respect to the weights. 

The Normalized LMS (NLMS) algorithm has been shown to achieve a better convergence rate compared to LMS by incorporating a different step-size parameter for each component $\mathbf{u}_i$ of the vector $\mathbf{u}$ \cite{adapt2}. Furthermore, LMS can encounter scalability issues with the input, choosing the step-size parameter sensitive to divergence. To address this, normalization is introduced to the update term: 
\begin{equation}
 \mathbf{w}(j) = \mathbf{w}(j-1) + \lambda \frac{\mathbf{e}(j)}{||\mathbf{u}(j)||^2_2} \mathbf{u}(j) 
 \label{eq:NLMS}
\end{equation}
and the NLMS converges to the Wiener filter solution of Eq.~(\ref{eq: linear estimation problem}) as long as $0<\lambda <2$. 

\begin{figure}[htbp]
\centering
  \centering
  \includegraphics[width=.8\linewidth]{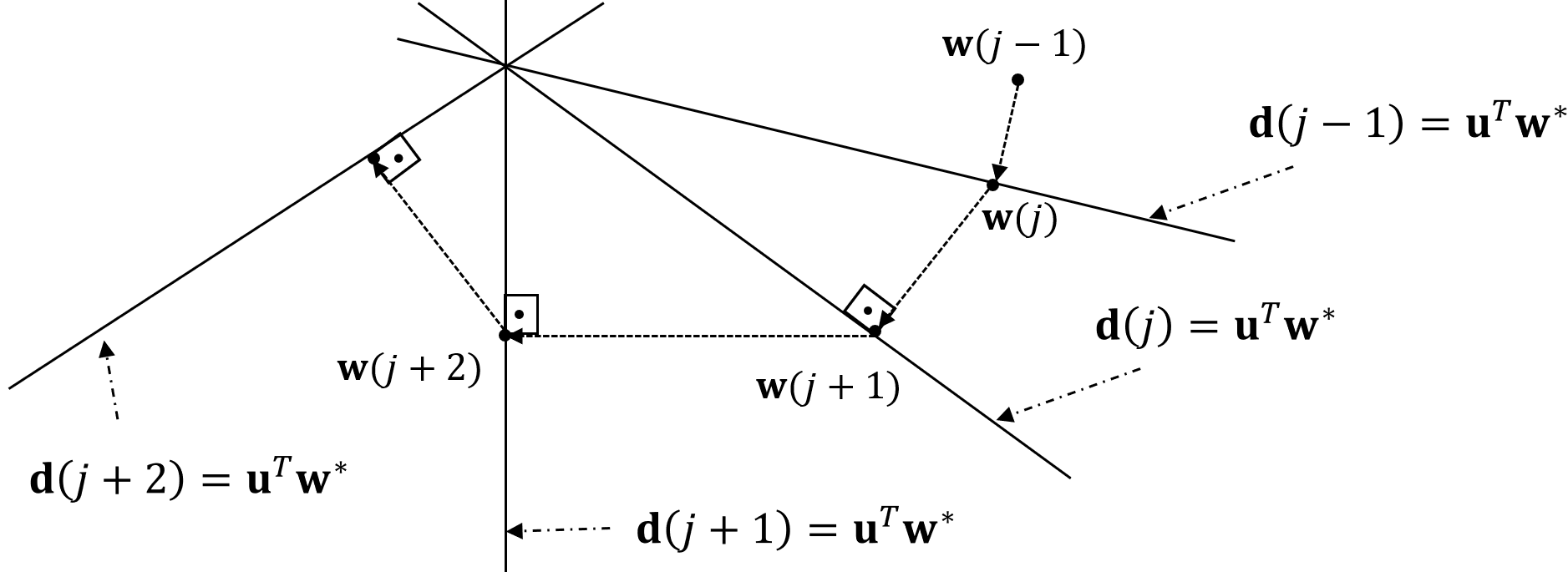}
\caption{NLMS projection.}
\label{fig:nlmsproj}
\end{figure}

 Another interpretation of the NLMS algorithm is based on the fact that the equation $\mathbf{e}(j)=\mathbf{d}(j) -\mathbf{u}^T\mathbf{w}$ is a hyperplane in the $M$ dimensional space $ \mathbf{w} \in \mathbb{R}^M$.
 When the vector $\mathbf{w}(j-1)$ is projected onto the hyperplane $\mathbf{e}(j)=\mathbf{d}(j) -\mathbf{u}^T\mathbf{w}$, we obtain the update equation
\begin{equation}
\mathbf{w}(j) = \mathbf{w}(j-1) +  \frac{\mathbf{e}(j)}{||\mathbf{u}(j)||^2_2} \mathbf{u}(j)
\end{equation}
as shown in Fig. \ref{fig:nlmsproj}. The orthogonal projection described in the above equation minimizes the Euclidean distance. 

Other distance measures lead to different update equations such as the $\ell_1$-norm-based updates:
\begin{equation}
\mathbf{w}(j) = \mathbf{w}(j-1) +  \frac{\mathbf{e}(j)}{||\mathbf{u}(j)||_1} \mathbf{u}(j)
\end{equation}
where $||\mathbf{u}(j)||_1$ is the $\ell_1$ norm of the vector $\mathbf{u}_j$ \cite{cetin1, kozat1, robust1, robust2, robust3}. The $\ell_1$-norm-based method is usually more robust to outliers in input.

In this article, we describe a new optimization algorithm inspired by Normalized LMS. It is called Input Normalized-SGD (INSGD) and utilizes the same approach as in NLMS. INSGD solves the constant learning rate issue that may cause divergence and obtains better accuracy results on benchmark datasets. The initialization of the learning rate or step size is crucial in DNN model training as it can greatly impact the convergence. We believe that incorporating the adaptive nature of NLMS into DNN training is a promising idea worth exploring. By adapting the concepts of NLMS to deep learning, we can potentially improve the convergence behavior and overall performance of DNN models.

\section{Methodology}
\subsection{Motivation} \label{sec:motivation}
Machine learning models commonly utilize the backpropagation method for optimization \cite{backprop, backprop2}. Stochastic Gradient Descent (SGD) is a widely used optimization algorithm with various modifications in the machine learning community. While SGD guarantees convergence with proper initialization, researchers have identified both positive and negative aspects of SGD and have attempted to enhance it according to their specific objectives \cite{ruder2016overview}.
One issue with SGD is that it updates weights based solely on the instantaneous gradient, which may lead to a lack of global information and oscillations. Another challenge is the use of a constant learning rate for all weights in the model. As training progresses, certain weights become more important than others, requiring different step sizes to ensure effective learning.

In recent years, the Adaptive Gradient (AdaGrad) algorithm was introduced by Duchi et al. as a means to enhance the update term in optimization. AdaGrad addresses the issue of choosing an appropriate learning rate by adapting it individually for each weight based on the cumulative sum of past and current squared gradients \cite{duchi2011adaptive}. By dividing the learning rate by the square root of this cumulative sum, AdaGrad assigns larger updates to weights with smaller gradients and vice versa. This adaptive approach allows for a more fine-grained adjustment of the learning rate based on the historical behavior of each weight's gradient. The formula for AdaGrad is  
\begin{equation}
\begin{split}
\mathbf{w}(k+1) \leftarrow \mathbf{w}(k) - \frac{\gamma}{\sqrt{v(k)+\epsilon}} \nabla_{\mathbf{w}(k)} L\\ 
v(k) \leftarrow v(k-1) + \left[\nabla_{\mathbf{w}(k)} L\right]^2
\end{split}
\end{equation}
where, $v$ is the weighted moving average of the squared gradient, $\gamma$ is the learning rate and $v(-1) = 0$. 

Hinton et al. introduced the RMSProp algorithm as an enhancement to the AdaGrad optimizer. RMSProp incorporates momentum by introducing an exponentially weighted moving average of the squared gradients. This modification helps to address the issue of diminishing learning rates in AdaGrad, which can slow down the convergence process. By applying the momentum concept, RMSProp allows for a smoother and more stable update process by considering not only the current squared gradients but also the historical information encapsulated in the moving average. As a result, RMSProp strikes a balance between the adaptability of AdaGrad and the stability provided by momentum, leading to improved optimization performance. RMSProp formula is
\begin{equation}
\begin{split}
\mathbf{w}(k+1) \leftarrow \mathbf{w}(k) - \frac{\gamma}{\sqrt{v(k)+\epsilon}} \nabla_{\mathbf{w}(k)} L\\ 
v(k) \leftarrow \beta v(k-1) + (1-\beta)\left[\nabla_{\mathbf{w}(k)} L\right]^2
\end{split}
\end{equation}

Another widely used optimization algorithm, called Adaptive Moment Estimation (Adam), was introduced by Kingma and Ba in 2014. Adam builds upon the concepts of momentum and the divisor factor used in RMSProp. In addition to maintaining an exponentially weighted moving average of the squared gradients like RMSProp, Adam also incorporates the notion of momentum by keeping track of an exponentially weighted moving average of the gradients themselves. This combination of momentum and the divisor factor makes Adam more adaptive and robust compared to RMSProp and AdaGrad. By considering both the first and second moments of the gradients, Adam adjusts the learning rate for each parameter individually, taking into account both the magnitude and direction of the gradients. This enables Adam to converge faster and handle a wider range of optimization scenarios. The algorithm is 
\begin{equation}
\begin{split}
\mathbf{w}(k+1) \leftarrow \mathbf{w}(k) - \frac{\gamma}{\sqrt{v(k)+\epsilon}} \mathbf{m}(k)\\ 
v(k) \leftarrow \beta v(k-1) + (1-\beta)\left[\nabla_{\mathbf{w}(k)} L\right]^2\\
\mathbf{m}(k) \leftarrow \beta \mathbf{m}(k-1) + (1-\beta)\nabla_{\mathbf{w}(k)} L
\end{split}
\end{equation}
where $\mathbf{m}$ is the momentum and $\mathbf{m}(-1)=\mathbf{0}$.

Another adaptive learning algorithm is proposed by Singh ~\textit{et al.}. In \cite{singh2015layer}, Singh ~\textit{et al.} presented layer spesific adaptive learning rate. According to \cite{singh2015layer}, the parameters in the same layer share similar gradients; therefore, the learning rate of the entire layer should be similar but different layers should have different learning rates. The work is described to adjust the learning rate to escape from the saddle points. Similar to the other algorithms, it uses $\ell_2$ norm in gradients.
\begin{equation} \label{lsalra}
\begin{split}
\mathbf{w}(k+1) \leftarrow \mathbf{w}(k) - \gamma (1 + log(1+1/||\nabla_{\mathbf{w}(k)} L||_2)) \nabla_{\mathbf{w}(k)} L\\ 
\end{split}
\end{equation}
The algorithm in (\ref{lsalra}) allows learning rate to become larger when the gradients are small. The aim is to correct the update term when the gradients are small in the high error low curvature saddle points. Therefore, the algorithm escapes from saddle points with large learning rate. Similarly, it scales the learning rate to stability if the gradients are too large. The use of $log$ function provides the scaling under different conditions.

Adam, AdaGrad, and RMSProp are optimization algorithms that address the limitations of standard stochastic gradient descent (SGD). These algorithms aim to improve the convergence speed in various scenarios, such as high learning rates or random weight initializations. While they incorporate normalization parameters, the update terms in these algorithms are still input-dependent and gradually decrease over iterations. In our approach, we propose using a normalization term based on the layer's input instead of relying on cumulative sums, which helps to prevent slow convergence. By leveraging input-based normalization, we aim to enhance the training process and overcome the limitations of existing optimization algorithms.

\subsection{Input Normalized Stochastic Gradient Descent (INSGD) Algorithm}
Input Normalized Stochastic Gradient Descent (INSGD) utilizes a similar approach as NLMS. The input scalability issue and the fragile nature of choosing the learning rate are the main issues that we address in the INSGD optimizer. 

In deep learning, we minimize the cost function
\begin{equation}
    F(\mathbf{W}) = \frac{1}{n} \sum_{k=1}^{n} F_k(\mathbf{W}) \notag
\end{equation}
where $\mathbf{W}$ represents the parameters of the network, $n$ is the number of training samples, and $F_k(\mathbf{W})$ is the loss due to the $k$-th training data. Let us first assume that we have linear neurons in the last layers of the network and $d_i$ is the desired value of the $i$-th neuron. Furthermore let $\mathbf{w_{i,0}}$ be the initial weights of the $i$-th neuron. We want the neuron to satisfy
\begin{equation}
    d_i = \mathbf{w} \cdot \mathbf{x} \notag
\end{equation}
where $\mathbf{x}$ denotes the input vector to the neuron. During training, we have $\mathbf{w_{i,0}} \cdot x_k \neq d_i$ where $\mathbf{x}_k$ is the input vector due to the $k$-th training pattern. We select the new set of weights of the neuron by solving 

\begin{equation} \label{eq:motive1}
    arg\min_{\mathbf{w}} ||\mathbf{w}_{i,0}-\mathbf{w}||^2  
\end{equation}
\begin{equation}
    st \: \mathbf{w} \cdot \mathbf{x}_k = d_i  \notag
\end{equation}
Using the Lagrangian multiplier, the solution to the optimization problem is the orthogonal projection onto the hyperplane $\mathbf{w} \cdot \mathbf{x}_k = d_i$. Solving (\ref{eq:motive1}) gives us an update equation
\begin{equation} \label{eq:noact}
    \mathbf{w}_{i,1} = \mathbf{w}_{i,0} + \lambda \frac{e_k}{\epsilon + ||\mathbf{x}_k||^2} \mathbf{x}_k
\end{equation}
where the update parameter $\lambda=1$, $\epsilon$ is a small number to avoid the division by zero and the error $e_k = d_i - \mathbf{w}_{i,0}\mathbf{x}_k$. This selection of weights is obviously reduces $F_k(\mathbf{W})$ and it is the same as the Gradient Descent with a new step size determined by the length of the input vector. It is also the well-known Normalized Least Mean Square (NLMS) algorithm used in adaptive filtering and signal processing as shown in Sec \ref{nlms}, Eq.~(\ref{eq:NLMS}). This equation is essentially the same as the NLMS equation Eq.~(\ref{eq: linear estimation update}). NLMS algorithm converges for $0 < \lambda < 2$ when the input is a wide-sense stationary random process. Inspired by the NLMS algorithm we can continue updating the neurons of the inner layers of the network in the same manner.

When the $i$-th neuron is not a linear neuron, we have
\begin{equation}
   \psi (\mathbf{w} \cdot \mathbf{x}) = d_i 
\end{equation}
where $\psi$ is the activation function. In this case, we solve the following problem to update the weights of the neuron.
\begin{equation} \label{eq:motive2}
    arg\min_{\mathbf{w}} ||\mathbf{w}_{i,0}-\mathbf{w}||^2  
\end{equation}
\begin{equation}
    st \: \psi(\mathbf{w} \cdot \mathbf{x}_k) = d_i  \notag
\end{equation}
or 
\begin{equation} \label{eq:motive3}
    arg\min_{\mathbf{w}} ||\mathbf{w}_{i,0}-\mathbf{w}||^2  
\end{equation}
\begin{equation}
    st \: \mathbf{w} \cdot \mathbf{x}_k = \phi(d_i)  \notag
\end{equation}
where $\phi$ is the inverse of the $\psi$ function. When $\psi$ is the sigmoid or tanh, $\psi$ has a well-defined inverse but since the inputs and outputs of any layer are available, we can infer the values numerically. In this case, the weight update equation will be
\begin{equation} \label{eq:motive_act}
    \mathbf{w}_{i,1} = \mathbf{w}_{i,0} + \lambda \frac{(\phi(d_i) - \mathbf{w}_{i,0} \cdot \mathbf{x}_k)}{\epsilon + ||\mathbf{x}_k||^2} \mathbf{x}_k
\end{equation}
By employing the solution described in Equation (\ref{eq:motive_act}), the NLMS algorithm can be adapted to optimize the weights in the final layer to minimize various cost functions. However, extending the INSGD algorithm to deeper networks with multiple layers poses a challenge in its derivation. We adopt similar assumptions to those used in the backpropagation algorithm to derive the INSGD algorithm for each weight in a deep-learning model. These assumptions provide a foundation for developing the INSGD algorithm, allowing us to effectively optimize the weights across the layers of the deep learning model.

\begin{figure*}[htbp]
    \centering
    \includegraphics[width=.8\linewidth]{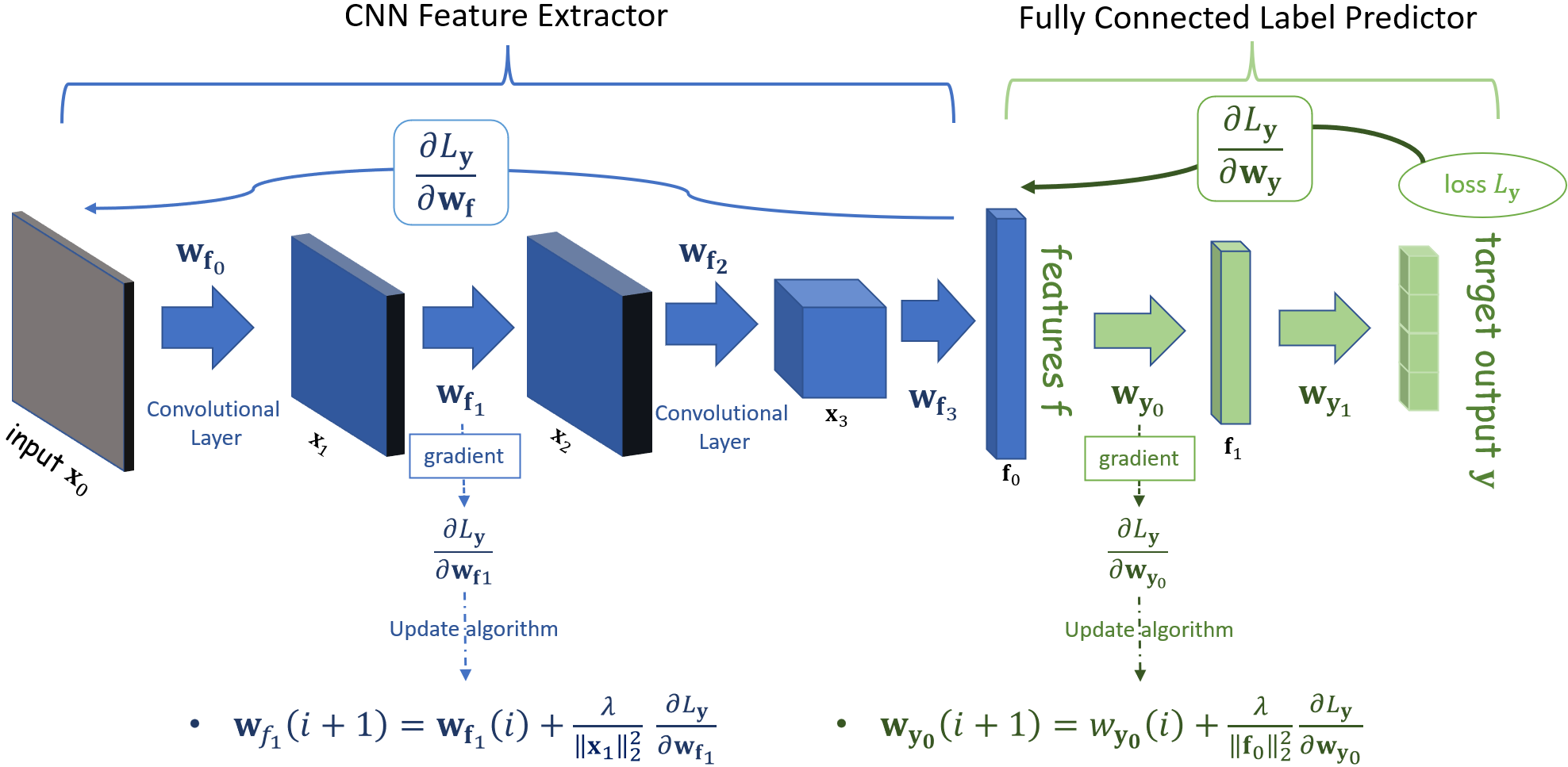}
    \caption{NSGD algorithm for different layers. It utilizes the input to each layer to update the weights.}
    \label{fig:backprop}
\end{figure*}

In addition to the final layer, we incorporate the input feature maps of each layer to apply the gradient term with normalization to the neurons using the backpropagation algorithm. This enables us to propagate the gradients and update the weights in a layer-wise manner throughout the network. By leveraging the information from the input feature maps, we enhance the training process by ensuring that the gradients are appropriately scaled and normalized at each layer. This approach allows for effective gradient propagation and weight updates, ultimately contributing to improved optimization and performance of the deep learning model. 

\begin{equation} \label{eq:NSGD2}
\mathbf{w}_{k+1} = \mathbf{w}_k - \mu \frac{\nabla_{\mathbf{w}_k}L(\mathbf{e}_k)}{\epsilon + ||\mathbf{x}_k||^2_2}
\end{equation}

where $\mathbf{x}_k$ is the vector of inputs to the neuron and $\mathbf{w}_k$ are the weights of the neurons. Note that we drop $i$ in the weight notation that represents the neuron since the algorithm is applicable to every neuron. For convenience, we also change the notation for the learning rate from $\lambda$ to $\mu$.
A description of  how the INSGD optimizer algorithm works for any layer of a typical deep network is shown in Fig. \ref{fig:backprop}.

The proposed INSGD algorithm, while addressing the input scalability problem, still shares some drawbacks with SGD. In Equation (\ref{eq:NSGD2}), we can observe how the weights are updated based on the input and gradient at each time step. However, the presence of irregular gradients and outlier inputs from the training dataset can impact the convergence behavior. To overcome this, we incorporate momentum, a technique that aids in navigating high error and low curvature regions \cite{ruder2016overview}. In the INSGD algorithm, we introduce an input momentum term to estimate the power of the dataset, enabling power normalization. By replacing the denominator term with the estimated input power, we emphasize the significance of power estimation in our algorithm. Furthermore, the utilization of input momentum allows us to capture the norm of all the inputs. Denoted as $P$, the input momentum term accumulates the squared $\ell_2$ norm of the input instances.

\begin{equation} \label{eq:input_momentum}
    P_k = \beta P_{k-1} + (1-\beta)||\mathbf{x}_{k}||_2^2 
\end{equation}

While estimating the input power is crucial, we encounter a challenge similar to AdaGrad. The normalization factor can grow excessively, resulting in infinitesimally small updates. To address this, we draw inspiration from the Layer Specific Adaptive Learning Rate approach and employ the logarithm function to stabilize the normalization factor. However, the use of the logarithm function introduces the risk of negative values. If the power is too low, the function could yield a negative value, reversing the direction of the update. To mitigate this, we employ a function with the rectified linear unit, which avoids the issue of negative values. Adding a regularizer may not be sufficient to resolve this problem, hence the choice of the rectified linear unit function. The function is designed as follows:

\begin{equation} \label{eq:f_clip}
f_{\epsilon}(u) = \left\{
	\begin{array}{ll}
		u  & \mbox{if } u \geq \epsilon \\
		\epsilon & \mbox{if } u < \epsilon
	\end{array}
\right.
\end{equation}

where $\epsilon$ is a regularizer to avoid the division by zero. After devising the function in (\ref{eq:f_clip}) and the logarithm approach in (\ref{eq:input_momentum}), the optimization algorithm for any weight in any layer in a network becomes 

\begin{equation} \label{eq:INSGD}
\mathbf{w}_{k+1} = \mathbf{w}_k - \frac{\mu}{f_{\epsilon}(log(P_k))} \nabla_{\mathbf{w}_k}L(\mathbf{e}_k)
\end{equation}

where $P_k$ is the estimate of the input power that is updated with every instance of $\mathbf{x}$ and the proposed $\epsilon = 0.01$. Therefore, we make sure that the update term is always greater than $0$ and stable. The iterative algorithm defined in Eq. (\ref{eq:INSGD}) is the Input Normalized SGD algorithm.

One can explore different norms, such as the $l_1$ or $l_{\text{max}}$ norm, as alternatives to the $l_2$ norm. In our experiments, we also investigated the use of the $l_1$ norm to assess its impact on performance. LMS algorithms based on the $l_1$ norm are known to be more robust against outliers in the input, which suggests potential benefits in deep neural network training. In this study, we examined both the $l_2$ and $l_1$ norms and their implications. Since NLMS is based on the $l_2$ norm, the algorithm presented in (\ref{eq:input_momentum}) utilizes the $l_2$ norm. However, for broader applicability, we can adapt the power estimation as follows:

\begin{equation}
    P_k^B = \beta P_{k-1}^B + (1-\beta)||\mathbf{x}_k^B||_p^p \notag
\end{equation}
where $||.||_p^p$ is the $p$ power of the $p$-norm, $p=1,2$, and the algorithm in (\ref{eq:input_momentum}) is the same. Extension of the NISGD to convolutional layers is straightforward. The pseudocode algorithm of NISGD is given in Alg. \ref{alg:insgd_m}.

\begin{algorithm}
\setstretch{1.2}
\caption{Input Normalized Gradient Descent with Momentum}\label{alg:insgd_m}
\begin{algorithmic}

\For{$t \gets 1$ to $...$}
    \State $g_t \gets \nabla_{\theta}f_t(\theta_{t-1})$ {\normalsize \Comment{Denote the gradient}}
        \If{$\beta \neq 0$}{\normalsize \Comment{If input momentum is not 0}}
            \If{$t > 1$}
                \State $P_t^B \gets \beta P_{t-1}^B + (1-\beta) ||\mathbf{x}_{t,\theta}^B||_2^2$ {\normalsize \Comment{Accumulate the power of input norm}}
            \Else
                \State $P_t^B \gets ||\mathbf{x}_{t,\theta}^B||_2^2$
            \EndIf
        \EndIf
        \State $g_t \gets$ {\Large $\frac{g_t}{f(log(P_t^B)}$} {\normalsize \Comment{Division by input norm}}
    \If{$\lambda \neq 0$} {\normalsize \Comment{Weight Decay}}
        \State $g_t \leftarrow g_t + \lambda\theta_{t-1}$ 
    \EndIf     
    \If{$\gamma \neq 0$} {\normalsize \Comment{Gradient with Momentum}}
        \If{$t > 1$}
            \State $\mathbf{b}_t \gets \gamma \mathbf{b}_{t-1} + (1-\tau) g_t$
        \Else
            \State $\mathbf{b}_t \gets g_t$
        \EndIf
        \State $g_t \gets \mathbf{b}_t$
    \EndIf
    \State $\theta_t \gets \theta_{t-1} - \mu g_t$ {\normalsize \Comment{Update the Weights}}
\EndFor

\end{algorithmic}
\end{algorithm}

\subsection{Models Architecture} \label{models}

In this study, we conducted experiments using five different networks to evaluate the performance of the INSGD algorithm in the classification tasks of CIFAR-10, CIFAR-100, and ImageNet-1K. We made modifications to the network architectures and initialization settings to assess the impact of the INSGD algorithm. In this study, we employed several networks for the classification tasks. Specifically, we utilized ResNet-20~\cite{he2016deep} for CIFAR-10, WResNet-18~\cite{wideresnet} for CIFAR-100, and ResNet-50 for ImageNet-1K. Additionally, we designed a custom CNN architecture specifically for CIFAR-10, which consists of four convolutional layers, each followed by a batch normalization layer. These networks were chosen to provide a diverse set of architectures and enable a comprehensive evaluation of the INSGD algorithm's performance across different datasets. As the NLMS algorithm is derived based on the mean squared error cost function, we employed both cross entropy loss and mean squared error loss as the objectives to be optimized in our experiments. The structure of ResNet-20 and custom-designed CNN used in this study is shown in Tables \ref{tab: resnet-20}, \ref{tab: toy}.

\begin{table}[htbp]

\centering
    \begin{tabular}{lcc}
    \hline\noalign{\smallskip}
		\bf{Layer}&\bf{Output Shape}&\bf{Implementation Details}\\
         \noalign{\smallskip}\hline\noalign{\smallskip}
		Conv1&$16\times32\times32$&$3\times3, 16$\\
		Conv2\_x&$16\times32\times32$&$\left[ \begin{array}{c} 3\times3, 16  \\  3\times3, 16 \end{array}\right]\times 3$\\
		Conv3\_x&$32\times16\times16$&$\left[ \begin{array}{c} 3\times3, 32  \\  3\times3, 32 \end{array}\right]\times 3$\\
		Conv4\_x&$64\times8\times8$&$\left[ \begin{array}{c} 3\times3, 32  \\  3\times3, 64 \end{array}\right]\times 3$\\
		GAP&$64$&Global Average Pooling\\
		Output&$10$&Linear\\
    \noalign{\smallskip}\hline\noalign{\smallskip}
\end{tabular}
\caption{ResNet-20 Structure for CIFAR-10 classification task. Building blocks are shown in brackets, with the numbers of blocks stacked. Downsampling is performed by Conv3\_1 and Conv4\_1 with a stride of 2. }
\label{tab: resnet-20}

\centering
    \begin{tabular}{lcc}
    \hline\noalign{\smallskip}
		\bf{Layer}&\bf{Output Shape}&\bf{Implementation Details}\\
         \noalign{\smallskip}\hline\noalign{\smallskip}
		Conv1&$8\times32\times32$&$3\times3, 8$\\
		Conv2&$16\times16\times16$&$3\times3, 16, \text{stride}=2$\\
		Conv3&$32\times8\times8$&$3\times3, 32, \text{stride}=2$\\
		Conv4&$64\times4\times4$&$3\times3, 64, \text{stride}=2$\\
		Dropout&$64\times4\times4$&$p=0.2$\\
		Flatten&$1024$&-\\
		Output&$10$&Linear\\
    \noalign{\smallskip}\hline\noalign{\smallskip}
\end{tabular}
\caption{Structure of the custom network with 4 conv layers for the CIFAR-10 classification task. }
\label{tab: toy}
\end{table}

On large benchmark datasets, traditional optimization algorithms often struggle to find the global optimum if the learning rate is not properly chosen. In such cases, these algorithms may diverge and fail to converge to the desired solution. However, the INSGD algorithm offers a solution by providing flexibility in learning rate selection, thereby improving the chances of reaching the global optimum. By adapting the learning rate dynamically based on the input and gradient information, INSGD enhances the optimization process and increases the likelihood of achieving superior results on large-scale datasets.

\section{Experimental Results}
Our experiments are carried out on an HP Workstation with an NVIDIA GeForce GTX 1660 Ti GPU for the CIFAR-10 and a HP Workstation with an NVIDIA RTX A6000 GPU for the CIFAR-100 and ImageNet-1K. 
Code is written in PyTorch in Python 3 and available at https://github.com/SalihFurkan/Normalized-SGD.
\subsection{CIFAR-10 Classification}
We conducted a series of experiments using the CIFAR-10 dataset which consists of 10 classes, initially employing the custom-designed CNN and ResNet-20 models for training. In certain experiments, we made modifications to the custom network to explore the algorithm's capabilities. These experiments aimed to assess the algorithm's performance under various conditions.

The base setting employed the SGD optimizer with a weight decay of 0.0005 and a momentum of 0.9. The models were trained using a mini-batch size of 100 for 200 epochs, with an initial learning rate ranging from 0.5 to 0.01. The learning rate was reduced at multiple steps with varying rates. To augment the data, we performed padding of 4 pixels on the training images, followed by random crops to obtain 32x32 images. Random horizontal flips were also applied to the images with a probability of 0.5. Normalization was performed on the images using a mean of [0.4914, 0.4822, 0.4465] and a standard deviation of [0.2023, 0.1994, 0.2010]. Throughout the training process, the best models were saved based on their accuracy on the CIFAR-10 test dataset. These settings were adopted from \cite{he2016deep}.


In the initial experiment, we employed the ResNet-20 model as our baseline. The independent parameter in this experiment was the learning rate, which we varied across different settings. The batch size is fixed at 128. We compared the accuracy results of our algorithm against those of other commonly used optimization algorithms, which are discussed in Section \ref{sec:motivation}. The detailed accuracy results can be found in Table \ref{tab: Compare_Opt}.

\begin{table}[htbp]
    \centering
    \begin{tabular}{cccc}
    \hline\noalign{\smallskip}
        \bf{Optimizer}& \bf{Initial Learning Rate}&  \bf{Test Accuracy} \\
     \noalign{\smallskip}\hline\noalign{\smallskip}
SGD  & 0.05 & 92.58\%  \\
SGD  & 0.1  & 92.55\%  \\
     \noalign{\smallskip}\hline\noalign{\smallskip}
Adam & 0.001 & 91.39\%  \\
     \noalign{\smallskip}\hline\noalign{\smallskip}
Adagrad & 0.01 & 87.29\% \\
Adagrad & 0.1 & 89.41\% \\
     \noalign{\smallskip}\hline\noalign{\smallskip}
Adadelta & 0.1 & 89.33\% \\
     \noalign{\smallskip}\hline\noalign{\smallskip}
INSGD-$\ell_1$  & 0.05 & 92.55\%  \\
INSGD-$\ell_1$  & 0.1 & 92.43\%  \\
     \noalign{\smallskip}\hline\noalign{\smallskip}
INSGD-$\ell_2$  & 0.05 & \textbf{92.56\%}  \\
INSGD-$\ell_2$  & 0.1 & \textbf{92.67\%}  \\
\noalign{\smallskip}\hline\noalign{\smallskip}
	\end{tabular}
\caption{Accuracy results of the ResNet-20 on the CIFAR-10 dataset with different initial learning rates using different optimization algorithms.
}
\label{tab: Compare_Opt}
\end{table}

The table clearly illustrates the significant improvement in accuracy achieved by the INSGD algorithm compared to other traditional optimization algorithms, resulting in better convergence during CIFAR-10 training. It is important to note that INSGD consistently performs at a high level across various initial learning rates. The superior performance of INSGD highlights its potential as a robust optimization algorithm for deep learning tasks, showcasing its effectiveness in addressing the challenge of tuning learning rates and achieving improved convergence.

In the second experiment, we explore the impact of varying batch sizes on the normalization factor to understand how input size affects the training process. Analyzing the results across different batch sizes is crucial due to the trade-off between time and memory usage. While larger datasets may benefit from larger batch sizes to expedite training time, it is important to consider the increased memory requirements. If our algorithm produces comparable results with larger batch sizes, it demonstrates its scalability. Table \ref{tab: Compare_Batch} presents the accuracy results of other algorithms and INSGD when training the model with different batch sizes. To accommodate the increased batch size, we adjust the learning rate according to the linear scaling rule described in \cite{goyal2017accurate}.

\begin{table}[htbp]
    \centering
    \begin{tabular}{cccc}
    \hline\noalign{\smallskip}
        \bf{Optimizer}& \bf{Batch Size}& \bf{Learning Rate}&  \bf{Test Accuracy} \\
     \noalign{\smallskip}\hline\noalign{\smallskip}
SGD  & 128 & 0.1 & 92.55\%  \\
NISGD-$\ell_1$  & 128 & 0.1 & 92.43\%  \\
NISGD-$\ell_2$  & 128 & 0.1 & \textbf{92.67\%}  \\
\noalign{\smallskip}\hline\noalign{\smallskip}
SGD  & 256 & 0.2 & 92.46\%  \\
NISGD-$\ell_1$  & 256 & 0.2 & 92.19\%  \\
NISGD-$\ell_2$  & 256 & 0.2 & \textbf{92.56\%}  \\
\noalign{\smallskip}\hline\noalign{\smallskip}
SGD  & 512 & 0.2 & 92.20\%  \\
NISGD-$\ell_1$  & 512 & 0.2 & \textbf{92.39\%}  \\
NISGD-$\ell_2$  & 512 & 0.2 & \textbf{92.80\%}  \\
\noalign{\smallskip}\hline\noalign{\smallskip}
	\end{tabular}
\caption{Accuracy results of the ResNet-20 on the CIFAR-10 dataset with different batch sizes.
}
\label{tab: Compare_Batch}
\end{table}

We also conducted experiments using the custom network for CIFAR-10 training to validate our algorithm. We employed similar settings to those used in ResNet-20. The accuracy results of the custom network with different initial learning rates are presented in Table \ref{tab: Compare_toy1}.

\begin{table}[htbp]
    \centering
    \begin{tabular}{cccc}
    \hline\noalign{\smallskip}
        \bf{Optimizer}& \bf{Initial Learning Rate}&  \bf{Test Accuracy} \\
     \noalign{\smallskip}\hline\noalign{\smallskip}
SGD  & 0.1 & 78.75\%  \\
NSGD-$\ell_1$  & 0.1 & 78.22\%  \\
NSGD-$\ell_2$  & 0.1 & 78.45\%  \\
\noalign{\smallskip}\hline\noalign{\smallskip}
SGD  & 0.25 & 58.95\%  \\
NSGD-$\ell_1$  & 0.25 & 78.77\%  \\
NSGD-$\ell_2$  & 0.25 & \textbf{78.96\%}  \\
\noalign{\smallskip}\hline\noalign{\smallskip}
SGD  & 0.03 & 77.43\%  \\
NSGD-$\ell_1$  & 0.03 & 78.11\%  \\
NSGD-$\ell_2$  & 0.03 & \textbf{79.14\%}  \\
\noalign{\smallskip}\hline\noalign{\smallskip}
	\end{tabular}
\caption{Accuracy results of the custom-designed CNN on the CIFAR-10 dataset with different initial learning rates and reduction rates.
}
\label{tab: Compare_toy1}
\end{table}

The toy network, used as a simplified representation of the model, plays a crucial role in evaluating the effectiveness of our algorithm. The results obtained from training the toy network validate the robustness of INSGD, as it consistently enhances the accuracy irrespective of the network architecture employed. Notably, when compared to SGD with momentum, INSGD consistently achieves superior performance across various learning rates, underscoring its efficacy in optimizing model training.
Given the overlap in the experiments conducted with the custom network and ResNet-20, we opted not to replicate the ResNet-20 experiments using the toy network. This decision was made to avoid redundancy in our findings and to focus on exploring the direct impact of NISGD

\subsection{CIFAR-100 Experiment}
We further extend our research by conducting experiments on the CIFAR-100 dataset. CIFAR-100 is a more challenging dataset compared to CIFAR-10 as it contains 100 classes instead of 10, requiring models to have a higher level of discrimination and classification capability. The increased class diversity in CIFAR-100 poses additional difficulty in achieving high accuracy and generalization performance. It is crucial to ensure that each class is adequately represented in the training process. Hence, we also opted to increase the batch size to 256 for this particular experiment. Before our study, Wide ResNet-18 was recognized for its convergence capabilities and satisfactory results \cite{wideresnet}. In alignment with the settings outlined in the Wide ResNet paper, we replaced the optimizer algorithm with INSGD. Similar to our CIFAR-10 experiment, the model was trained for 200 epochs, and we report the highest accuracy achieved on the testing data. 

\begin{table}[htbp]
    \centering
    \begin{tabular}{ccccc}
    \hline\noalign{\smallskip}
        \bf{Optimizer}& \bf{Initial Learning Rate}& \bf{Batch Size} &  \bf{Top-1 Accuracy} &  \bf{Top-5 Accuracy}\\
     \noalign{\smallskip}\hline\noalign{\smallskip}
SGD  & 0.1 & 128 & 78.75\% & 94.20\%  \\
NSGD-$\ell_1$  & 0.1 & 128 & 78.52\% & \textbf{94.66\%}  \\
NSGD-$\ell_2$  & 0.1 & 128 & \textbf{78.85\%} & \textbf{94.34\%}  \\
     \noalign{\smallskip}\hline\noalign{\smallskip}
SGD  & 0.1 & 256 & 77.22\% & 93.79\%  \\
NSGD-$\ell_1$  & 0.1 & 256 & \textbf{78.15\%} & \textbf{94.54}\%  \\
NSGD-$\ell_2$  & 0.1 & 256 & \textbf{77.89\%} & \textbf{93.98\%}  \\
\noalign{\smallskip}\hline\noalign{\smallskip}
	\end{tabular}
\caption{Accuracy results of the Wide ResNet-18 on the CIFAR-100 dataset.
}
\label{tab: Compare1}
\end{table}

The results presented in Table \ref{tab: Compare1} provide compelling evidence of the effectiveness of the INSGD algorithm in achieving improved convergence on complex datasets across a range of learning rates. The superior performance of INSGD, as evidenced by its higher Top-1 and Top-5 accuracy, establishes its utility in training sophisticated models on challenging datasets. These findings underscore the algorithm's capability to handle intricate data distributions and optimize model performance, thereby showcasing its potential for advancing the state-of-the-art in deep learning.

\subsection{ImageNet-1K Results}

In this subsection, we present the test accuracy results for the ImageNet-1K dataset. We utilize the ResNet-50 model, as discussed in Section \ref{models}. The training process is conducted using the official PyTorch ImageNet-1K training code \cite{ImageNet_training_in_PyTorch}. Specifically, we employ the SGD and NISGD optimizers with a weight decay of 0.0001 and a momentum of 0.9.

The ImageNet-1K dataset consists of 1.2 million images and is known for its difficulty in training. Due to the image resolution and resource constraints, adopting larger batch sizes is not feasible in our environment. As a result, we train the models with a mini-batch size of 256, an initial learning rate of 0.1 for 90 epochs, and a learning rate reduction of 1/10 after every 30 epochs.

To augment the data, we perform random cropping and horizontal flipping with a probability of 0.5, resulting in 224 × 224 images. The images are then normalized using a mean of [0.485, 0.456, 0.406] and a standard deviation of [0.229, 0.224, 0.225].

The accuracy of the best models is presented in Table \ref{tab: ImageNet-1K}, based on the center-crop top-1 accuracy and top-5 accuracy on the ImageNet-1K validation dataset. These accuracies are obtained from the model with the highest center-crop top-1 accuracy, providing a comprehensive evaluation of the model's performance on the ImageNet-1K dataset.

\begin{table}[htbp]
    \centering
    \begin{tabular}{cccc}
    \hline\noalign{\smallskip}
        \bf{Optimizer}& \bf{Initial Learning Rate}&  \bf{Top-1 Accuracy} &  \bf{Top-5 Accuracy}\\
     \noalign{\smallskip}\hline\noalign{\smallskip}
SGD  & 0.05 & 75.20\% & 92.49\%  \\
SGD  & 0.1 & 75.56\% & 92.53\%  \\
     \noalign{\smallskip}\hline\noalign{\smallskip}
NSGD-$\ell_1$  & 0.05 & \textbf{75.59\%} & \textbf{92.74\%}  \\
NSGD-$\ell_1$  & 0.1 & ??\% & ??\%  \\
     \noalign{\smallskip}\hline\noalign{\smallskip}
NSGD-$\ell_2$  & 0.05 & \textbf{75.67\%} & \textbf{92.66\%}  \\
NSGD-$\ell_2$  & 0.1 & \textbf{75.79\%} & \textbf{92.81\%}  \\
\noalign{\smallskip}\hline\noalign{\smallskip}
	\end{tabular}
\caption{Accuracy results of the ResNet-50 on the ImageNet-1K dataset.
}
\label{tab: ImageNet-1K}
\end{table}

The results presented in Table \ref{tab: ImageNet-1K} highlight the improved top-1 accuracy achieved by the INSGD algorithm on the ImageNet-1K dataset. This improvement is particularly significant considering the scale of the dataset, demonstrating the effectiveness of INSGD in handling large and complex datasets. By leveraging the input normalization factor, INSGD enables the model to converge more effectively by aligning the gradient direction and appropriate magnitude.

The power estimation obtained through momentum in INSGD indicates that the optimization algorithm can benefit from considering the entire input sequence. It suggests that the algorithm can capture long-term dependencies and utilize them for better optimization performance. Furthermore, it is worth noting that the batch size used in our experiments is relatively small compared to the number of images in the dataset. Exploring the algorithm's behavior with larger batch sizes would be an interesting avenue for future investigation.

\section{Conclusion}
In this paper, we proposed a novel normalization scheme called INSGD, which incorporates ideas from the widely used NLMS algorithm in adaptive filtering. INSGD introduces a normalization step that focuses on the input vector to the neurons, utilizing both the $l_1$ and $l_2$ norms.

To evaluate the effectiveness of INSGD, we conducted experiments on various datasets using different models. Notably, our algorithm consistently demonstrated improvements in testing accuracy across multiple datasets. For example, on the CIFAR-10 dataset, INSGD achieved a significant boost in accuracy compared to traditional stochastic gradient algorithms. We observed similar positive outcomes on other datasets, such as CIFAR-100 and ImageNet-1K, when employing different models like ResNet-20 and ResNet-50.

The promising results obtained across diverse datasets and models validate the effectiveness of NSGD in enhancing the training process. By incorporating the normalization factor into the stochastic gradient algorithm, INSGD effectively leverages the benefits of the NLMS algorithm, leading to improved performance in various scenarios.

\bibliographystyle{IEEEtran}
\bibliography{main}
\end{document}